\documentclass[10pt,final,journal]{IEEEtran}

\usepackage{amsmath,amssymb,amsfonts}
\usepackage{graphicx}
\usepackage{textcomp}
\usepackage{algorithm}
\usepackage[noend]{algpseudocode}
\usepackage{setspace}
\usepackage{enumitem}

\usepackage{tabularx}
\usepackage{multirow}
\usepackage{multicol}
\usepackage{diagbox}
\usepackage{adjustbox}
\usepackage{subcaption}
\usepackage[colorlinks,linkcolor=blue]{hyperref}
\usepackage[numbers,sort&compress]{natbib}
\usepackage{xcolor}
\usepackage{adjustbox}
\usepackage{booktabs}

\begin{document}

\title{Countermeasures Against Adversarial Examples in Radio Signal Classification}

\author{Lu Zhang$^{1}$, Sangarapillai Lambotharan$^{1}$, Gan Zheng$^{1}$, Basil AsSadhan$^{2}$, Fabio Roli$^{3}$ \\$^{1}$Wolfson School of Mechanical, Electrical and Manufacturing Engineering, Loughborough University, Loughborough, UK\\$^{2}$Department of Computer Science, King Saud University, Riyadh, Saudi Arabia\\$^{3}$Dept. of Electrical and Electronic Engineering, University of Cagliari, Piazza d’Armi, 09123 Cagliari, Italy}

\IEEEtitleabstractindextext{
\begin{abstract}
Deep learning algorithms have been shown to be powerful in many communication network design problems, including that in automatic modulation classification. However, they are vulnerable to carefully crafted attacks called adversarial examples. Hence, the reliance of wireless networks on deep learning algorithms poses a serious threat to the security and operation of wireless networks. In this letter, we propose for the first time a countermeasure against adversarial examples in modulation classification. Our countermeasure is based on a neural rejection technique, augmented by label smoothing and Gaussian noise injection, that allows to detect and reject adversarial examples with high accuracy. Our results demonstrate that the proposed countermeasure can protect deep-learning based modulation classification systems against adversarial examples.

\end{abstract}

\begin{IEEEkeywords}
deep learning, adversarial examples, radio modulation classification, neural rejection, label smoothing
\end{IEEEkeywords}}

\maketitle

\IEEEdisplaynontitleabstractindextext
\IEEEpeerreviewmaketitle

\section{Introduction}

Automatic modulation classification (AMC) is widely used in both civilian and military communications. For example, in civilian applications, it is useful for adaptive modulation and coding schemes, where the modulation and coding will be changed according to the instantaneous signal to noise ratio (SNR) and fading. Identification of modulation at the receiver can simplify the overhead bandwidth required for informing the receiver of any changes in modulations. AMC is also used for detection of primary users in cognitive radio networks, where the aim is to sense active transmitters and if there is none, the corresponding frequency can be used by an opportunistic transmitter \cite{6336689}. In addition, AMC is very important in military applications for automatic discovery of modulations used by the adversaries \cite{Swami2000}. 

Historically, radio signal classification relies on carefully hand-crafting signal features which requires expert knowledge and experience. As proven to be successful in many other application domains, deep learning has recently been applied to radio signal classification \cite{OShea2018,Scholl2019}. Despite of its potential benefits, previous work \cite{goodfellow2014explaining} has shown that the deep neural networks (DNNs) are highly vulnerable to adversarial examples (attacks), where the original inputs are perturbed with imperceptible and carefully designed modifications which can mislead classification outputs of DNNs. 

The effect of adversarial examples in modulation classification can be viewed differently for the civilian and military scenarios. In the civilian setup, adversarial perturbation can be viewed as a jamming strategy where an eavesdropper closer to the transmitter could decode the signals and transmit adversarial perturbations on the fly, assuming that the transmitter would continue with the same modulation type for a reasonable amount of time, i.e., spanning multiple symbols. At the legitimate receiver, the legitimate transmitted signals will be corrupted by the addition of perturbation signals which will lead to misclassifications at the receiver. In the military domain, one may wish to eavesdrop the communications of adversaries. Hence automatic discovery of modulation is required at the receiver. In this case, knowing the possibility of eavesdroppers, the transmitter of the adversaries could add a small amount of perturbations so that AMC at the receiver will fail. The defense technique proposed in this letter is applicable to both scenarios, however, in order to explain the concepts, we focus only on one scenario, the military one.

\begin{figure}[ht]
\centering
\includegraphics[scale = 0.3]{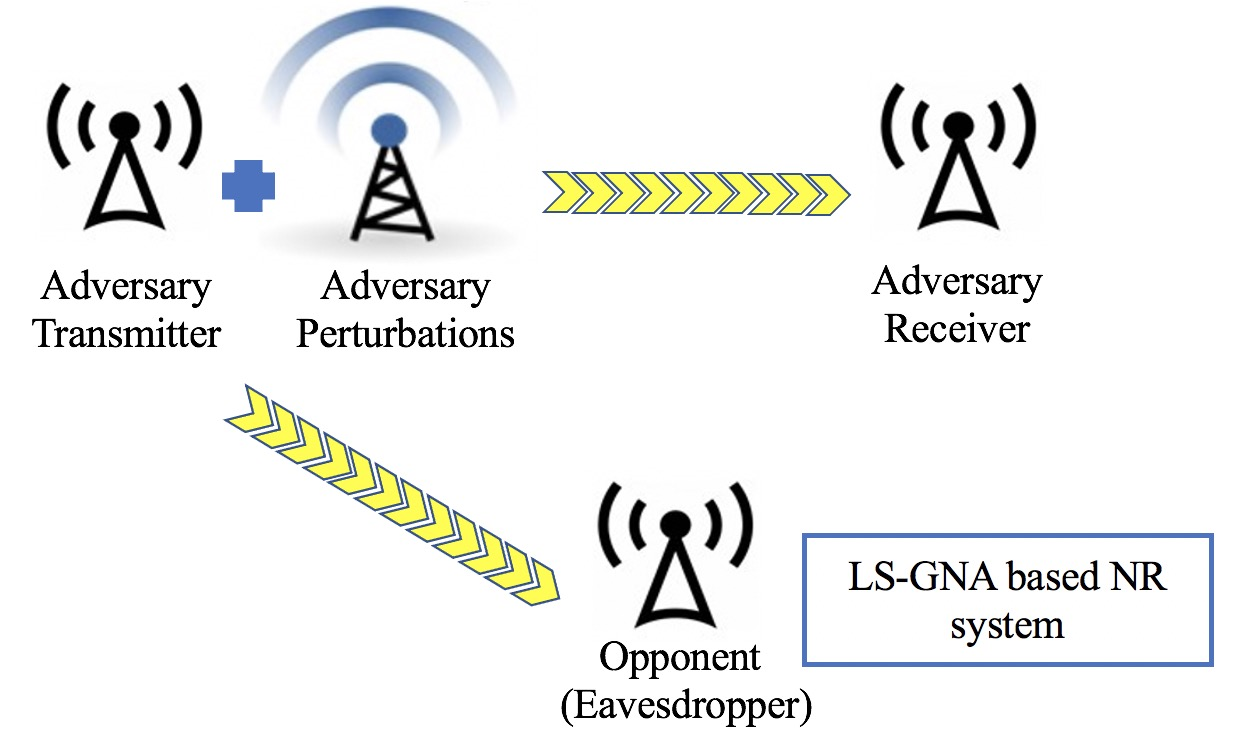}
  \caption{A military scenario for the adversarial examples in modulation classification.}
  \label{fig:eavesdropper}
\end{figure}

For the military scenario, surveillance and threat analysis in electronic warfare monitors the radio transmission channels for communication between the adversary’s units as shown in Figure \ref{fig:eavesdropper}. Once a communication activity is detected in a certain frequency band, modulation classification becomes one of the preliminary processes required for the recovery of the transmitted signal. Hence, the adversary will try to increase the difficulty of modulation classification. In short, in this military scenario, when an adversary transmits signals to its own receiver, the opponent (military/defender) aims to apply AMC and decode the signals. In order to deceive the opponent, the adversary will add a small amount of perturbation, hence the automatic discovery of the modulation by the opponent receiver will be unsuccessful. Hence, a countermeasure using a neural rejection (NR) system is employed at the receiver of the opponent to defend against adversarial perturbations. In the absence of a defense based on NR, every misclassifications will lead to decoding failure. However, in the presence of a defense based on NR, the aim of the opponent is to either correctly classify the modulation or to reject adversarial examples. In the case of rejection of adversarial examples, the opponent will be able to recognize the existence of adversarial transmissions and the modulation classification will be deemed unreliable. Therefore, the opponent can avoid wasting computational resources with futile attempts of signal decoding.

\setlength{\textfloatsep}{0.6mm}

In this work, we consider a worst-case attack scenario for the defender, i.e. all attacks correspond to the white-box attacks \cite{Sotgiu2020}. To be specific, there are three assumptions for the white-box attack. First, the adversary knows the exact input. Second, the adversary is synchronous with the transmitter such that each element of the input is perturbed with its corresponding element of the carefully-designed adversarial perturbation. These two conditions are satisfied in this threat scenario as the adversary is the transmitter itself. Third, the adversary should know the architecture of the opponent's classifier, which could come from separate intelligence.

Previous work has shown that the deep learning (DL) based modulation classification is extremely vulnerable to adversarial examples generated using fast gradient method (FGM) \cite{Sadeghi2019}. The perturbation is considered imperceptible if the perturbation-to-noise ratio (PNR) is less than one as the perturbation is of the same order as or even below the noise level \cite{Sadeghi2019}. However, to the best of our knowledge, no defense mechanism for adversarial attacks against modulation classification has been proposed so far. We propose for the first time a defense technique against adversarial attacks in radio signal classification. To make clear our contribution to the state of the art and for a fair comparison, we use the same dataset and the same DNN as in \cite{Sadeghi2019}. In this work, we show that an NR scheme \cite{Sotgiu2020} can protect the modulation classification system considerably as compared to the performance of an undefended DNN against adversarial attacks. To enhance the defense further by taking advantage of the robustness of label smoothing (LS) and Gaussian noise augmentation (GNA) \cite{Shafahi2019}, we propose a new NR system using LS and GNA, denoted as LS-GNA based NR system. The main contributions are as follows:

\begin{itemize}[leftmargin=0.5cm]
\item We are the first to propose a countermeasure against adversarial attacks in radio signal classification;
\item We propose a new defense technique using a rejection technique augmented by label smoothing and Gaussian noise injection. We demonstrate through experiments that the proposed LS-GNA based NR achieves a high classification accuracy against adversarial examples and outperforms both undefended DNNs and defense techniques that do not use label smoothing and Gaussian noise injection.
\end{itemize}


\section{THE PROPOSED COUNTERMEASURE AGAINST ADVERSARIAL ATTACKS}

\subsection{DNN against FGM Attacks}
We first illustrate the undefended DNN against FGM attack, which is considered as a benchmark. We used the same deep convolutional neural network (CNN) model VT-CNN2 as that used in \cite{Sadeghi2019}. To train the CNN classifier, we randomly choose half of the 110000 data samples as the training set and the rest as the test set. After training, to generate adversarial examples to evaluate the robustness of undefended VT-CNN2 classifier against FGM attack, we randomly choose 1000 samples from the test dataset corresponding to an SNR of 10dB. We separate these 1000 samples into two sets called set \uppercase\expandafter{\romannumeral1} and set \uppercase\expandafter{\romannumeral2}, that correspond to the samples that are correctly classified and misclassified respectively by the trained VT-CNN2 in the absence of adversarial perturbations. Then we generate adversarial examples separately for set \uppercase\expandafter{\romannumeral1} and set \uppercase\expandafter{\romannumeral2}. The rationale is that the data set \uppercase\expandafter{\romannumeral2} has difficult samples which are misclassified even in the absence of adversarial perturbations, while data set \uppercase\expandafter{\romannumeral1} consists of data that are correctly classified. Therefore, it is important to analyze independently these two data sets, and to evaluate the capability of the proposed countermeasure to detect adversarial manipulations of samples from the good data set \uppercase\expandafter{\romannumeral1}.



The FGM attack in our work is adopted from \cite{Sadeghi2019}. For all the possible targeted classes, we first calculate the gradient of the loss function $\triangledown _{x}L(x, e_{class\text{-}index})$, where $L(\cdot)$ means the cross entropy loss of the predicted output and the targeted class. Let $r_{norm}$ denote the normalized gradient. The perturbation $r_{x}$ is calculated by multiplying $r_{norm}$ by the perturbation norm $\varepsilon$. For every data sample, the perturbation norm $\varepsilon$ should be chosen as the smallest $\varepsilon$ that leads to misclassification. This can be obtained using a line search or using a computationally efficient bisection method. For a given $\varepsilon$, the corresponding PNR is $\varepsilon ^2 (SNR+1)/\left \| x \right \|_2^{2}$. Hence, for the analysis of performance in terms of accuracy against PNR, we can set $\varepsilon$ as $\varepsilon =\sqrt{PNR\cdot \left \| x \right \|_{2}^{2}/(SNR+1)}$ for a required PNR. Finally, the adversarial example for a specific targeted class is obtained by adding the perturbation $r_{x}$ to the original input. 






\subsection{Neural Rejection against FGM Attacks}
We propose to employ a robust defense called NR system for modulation classification. The NR system can detect the adversarial attacks through a simple rejection mechanism \cite{Sotgiu2020}. For a given data sample $x$, the NR system extracts the last layer outputs of a pre-trained DNN as features and feeds them into a one-versus-all support vector machine (SVM) classifier to generate decision scores for all the possible classes. If the maximum of decision scores associated with all the classes is less than a specific threshold $\Theta$, the data sample $x$ would be deemed an adversarial attack and will be rejected.

\begin{figure}[ht]
\centering
\includegraphics[scale = 0.52]{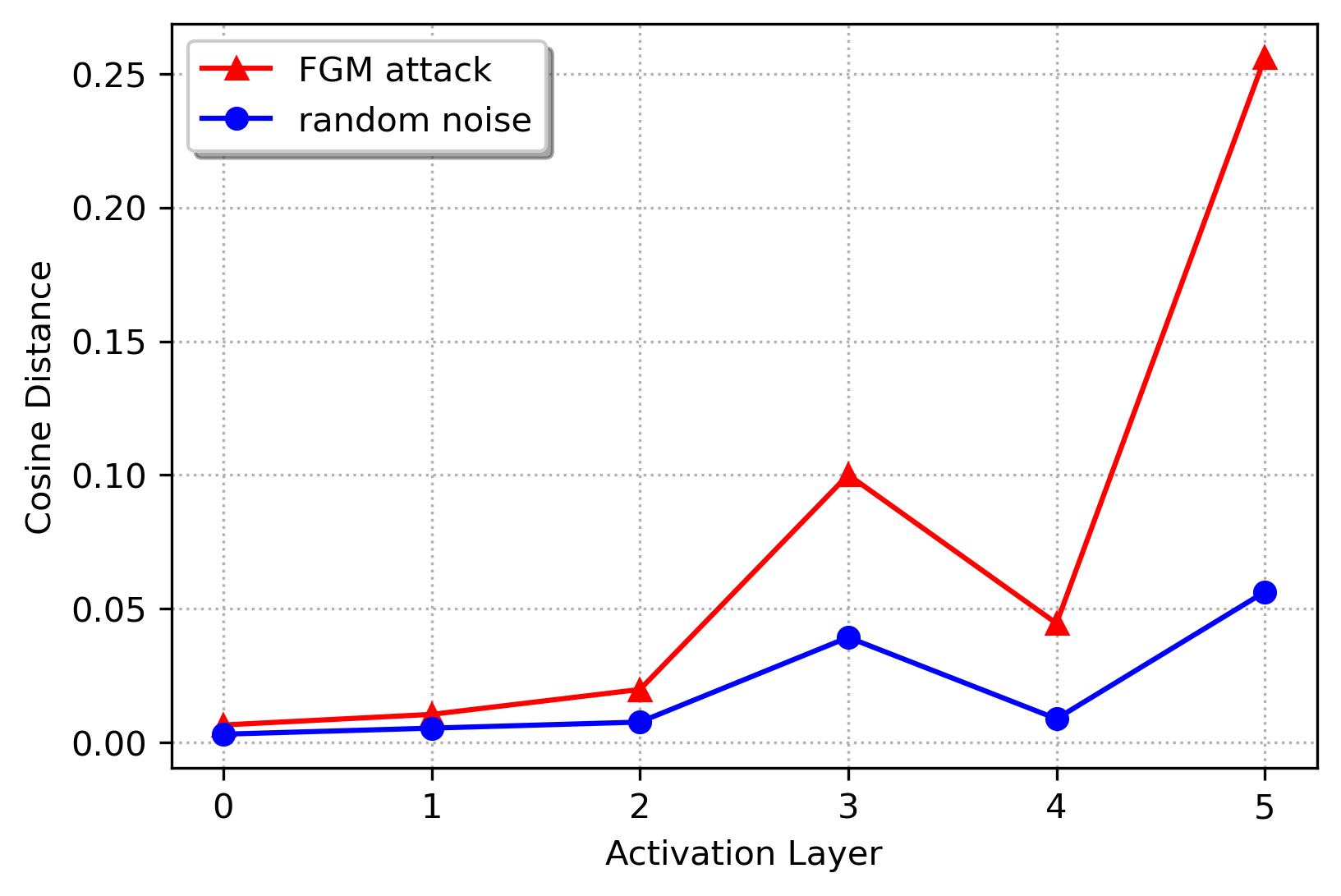}
  \caption{Conceptual explanation of the “amplification” phenomenon of adversarial examples in deep neural networks.}
  \label{fig:amplification}
\end{figure}

The rationale behind the rejection of adversarial examples is based on the fact that, for adversarial examples, the values of outputs of neurons of DNN become larger and larger during the propagation over the layers and the values are much larger than the values of the outputs related to the “normal” samples for the last layers. This phenomenon is known as “amplification” for the outputs of neurons of DNN during the “propagation” of adversarial examples through the layers of DNN. This “amplification” depicted in Figure \ref{fig:amplification} does not exist for “normal” samples even if they are noisy. To demonstrate this, we calculated the cosine distance between the samples at each activation layer, when the input to the DNN was the normal signal $x$ and the attack signal $\widetilde{x}$. The average of the cosine distance was obtained using all the test samples corresponding to SNR=10dB. The cosine distance indicates the dissimilarity between $x$ and $\widetilde{x}$, and is calculated as $1-\frac{x\cdot \widetilde{x}}{\left \| x \right \|\cdot \left \| \widetilde{x} \right \|}$ where $x\cdot \widetilde{x}$ is the inner product between the vectors $x$ and $\widetilde{x}$. For comparison, we also calculate the cosine distance between normal and noisy signals for each layer, where noisy signals indicate the normal samples augmented with random noise. Both FGM attacks and noisy signals have the same amount of perturbation. As shown in Figure \ref{fig:amplification} for the modulation classification, the dissimilarity between FGM attacks and normal signals becomes progressively larger during the propagation over the layers of DNN. The differences between FGM attacks and benign samples are much larger than that related to the noisy signals for the last layer. In other words, there is a sort of “amplification” for the difference between FGM attacks and benign samples during the “propagation” of adversarial examples through the layers of DNN. This amplification phenomenon is significant at the last layer, hence it is evident that anomalies due to adversarial perturbation of the radio signal can be detected and rejected at the last layer DNN output. Hence, our neural rejection technique works using the input from the last layer of DNN.

\vspace{-3mm}

\subsection{LS-GNA based NR System against FGM Attacks}
To further enhance our defense, we propose a new LS-GNA based NR system which obliges the adversary to use much more perturbation (transmission) power to fool the defender. The underlying concept of label smoothing and GNA stems from the fact that overconfidence in prediction can yield the DNN to be sensitive to adversarial perturbations. Hence uncertainty is introduced deliberately to the labels of the training dataset in the form of label smoothing and to the training data samples by GNA. This helps avoiding overfitting, but also helps the DNN and the NR system to be robust to adversarial attacks. LS is traditionally used to prevent over-fitting in general classification problems \cite{Shafahi2019}. It converts one-hot encoded label vectors into smoothed vectors that represent a low-confidence classification \cite{Shafahi2019}. Specifically, instead of using one-hot encoded labels $l$ to train the CNN, smoothed labels $\tilde{l}$ as $\tilde{l}= l-\alpha \times (l-\frac{1}{N_{c}})$ are used, where $N_{c}$ is the number of classes, and $\alpha \in [0,1]$ is the smoothing parameter. In this work, we chose $\alpha =0.1$. We provide an example to illustrate the LS technique. Assume that the one-hot label for the modulation QPSK is (1, 0, 0, 0, 0, 0, 0, 0, 0, 0, 0), then the smoothed label for the LS scheme can be (0.9091, 0.0091, 0.0091, 0.0091, 0.0091, 0.0091, 0.0091, 0.0091, 0.0091, 0.0091, 0.0091). LS helps DNN to have a better generalization and not being over-confident. A criterion for analyzing the robustness of defense against adversarial perturbation is the ratio of the logits difference to the gradient difference \cite{Shafahi2019}. For the proposed NR system, this can be written as,
\begin{equation}
\label{equ:robustness of NR}
\varepsilon_{L} > \frac{S_{y}(x)-S_{\bar{y}}(x)}{\left \| \triangledown _{x}S_{y}(x)-\triangledown _{x}S_{\bar{y}}(x) \right \|_{1}}
\end{equation}
where $S_{y}(x)$ and $S_{\bar{y}}(x)$ are the SVM outputs of the NR system corresponding to the true class $y$ and the nearest competing false class $\bar{y}$, respectively. $\triangledown _{x}S_{y}(x)$ and $\triangledown _{x}S_{\bar{y}}(x)$ are the gradients of $S_{y}(x)$ and $S_{\bar{y}}(x)$ with respect to the input $x$, respectively. The numerator quantifies the distance between the true class and the nearest competing false class. The denominator characterizes the sensitivity of the perturbation to the output of the SVM of the NR system. Hence, a good defense should maximize the inter-class distance and minimize the sensitivity of the perturbation to the output of the NR based classifier. Using the GNU radio dataset used in the simulation section, we obtained the mean of $\varepsilon_{L}$ for the NR system with and without LS-GNA as 0.00167 and 0.00113. Hence, LS-GNA is able to enhance the robustness of NR system against adversarial perturbations.

\begin{figure}[ht]
\centering
\includegraphics[width=\columnwidth]{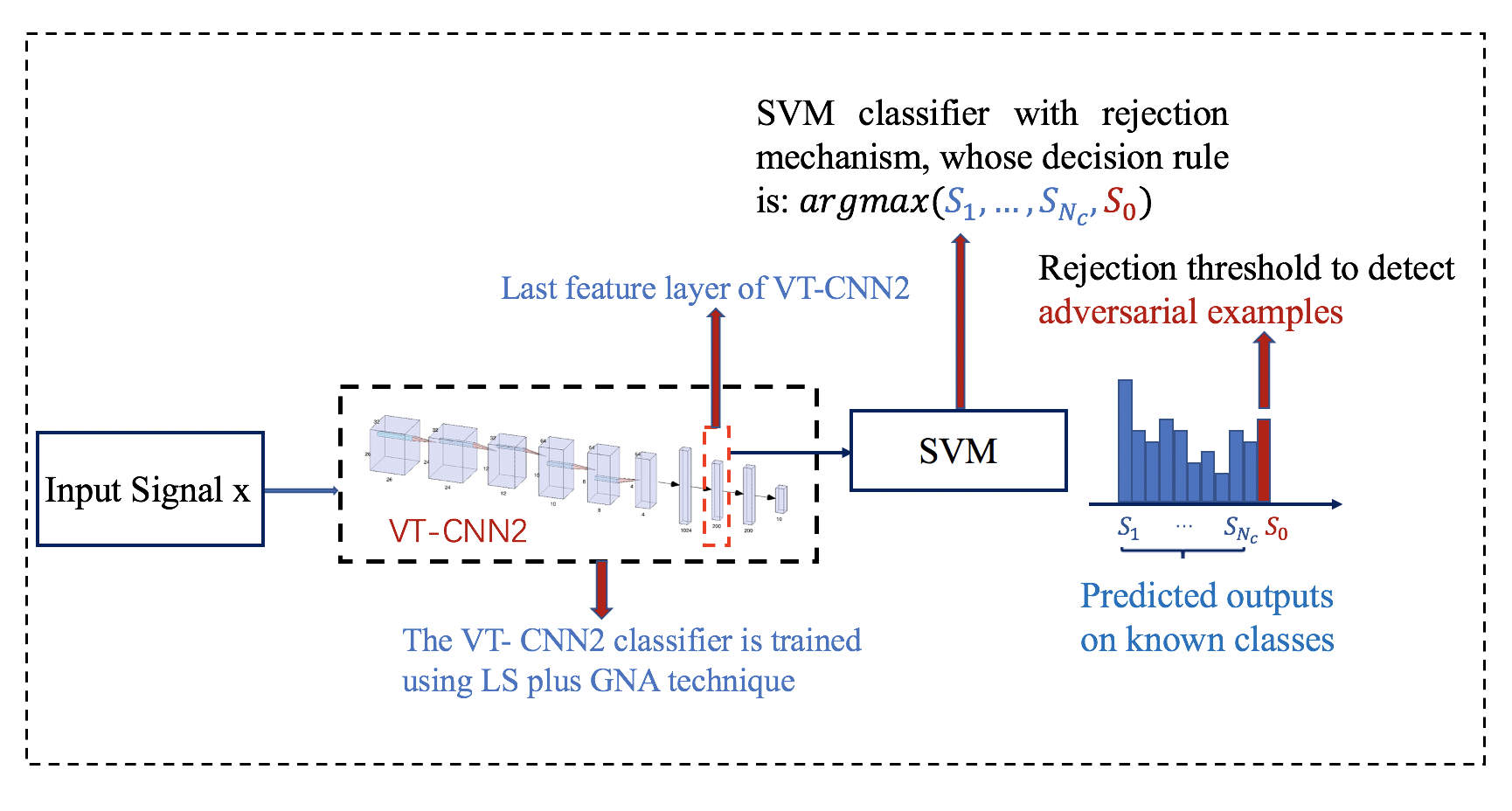}
  \caption{The architecture of the LS-GNA based NR system. The outputs of the last layer of the VT-CNN2 are extracted as features and fed into the SVM classifier that can detect and reject adversarial examples.}
  \label{fig:NR}
\end{figure}

The architecture of the LS-GNA based NR system is shown in Figure \ref{fig:NR}. As the first stage of the LS-GNA based NR system, the VT-CNN2 classifier is trained using LS and GNA. For GNA, we added zero mean Gaussian random noise into the data samples during the training of CNN. The variance of the Gaussian noise is $0.003$. After training the VT-CNN2 classifier, we randomly choose 10000 samples from the training set and extract the last layer features of these samples from the pre-trained CNN. These features were used to train the one-vs-all SVM classifier. The three-fold cross-validation was adopted to choose the hyper-parameters of the SVM classifier. During the testing phase, given an input signal $x$, the outputs of the last feature layer from VT-CNN2 classifier are extracted as features $f(x)$ and fed into the connected SVM classifier as shown in Figure \ref{fig:NR}. Then the predicted decision scores $S_{1}(x),...S_{N_{c}}(x)$ (i.e., the predicted outputs from SVM classifier) for all the possible classes were generated. The NR will make a decision in favour of class $c^{*}$ according to the decision function in (2).

\begin{equation}
\label{equ:decision_function}
c^*=arg \max_{k=1,...,N_{c}}S_{k}(x), ~only ~if~S_{c^{*}}(x)>\Theta
\end{equation}

The input signal $x$ is considered as an adversarial example if the maximum of the decision scores is less than the predefined rejection threshold $\Theta$. The  algorithm for generating FGM attacks for the LS-GNA based NR system is shown in Algorithm \ref{alg:FGM}. Compared to generating FGM attacks for the undefended DNN, there are two main differences in Algorithm \ref{alg:FGM}. The first one is the stopping criterion. In addition to the condition that the predicted label of the generated adversarial example is unequal to the true label, we have another condition that the maximum of the decision score of the crafted adversarial example is larger than the rejection threshold $\Theta$ as in line 5. The second difference is that the objective function $L\left ( x, \cdot  \right )$ is expressed as $ S_{y}(x)-S_{t}(x)$, where $y$ means the true class and $t$ means the targeted wrong class. Therefore, given an input signal $x$, the attacker attempts to minimize the confidence score of $x$ which belongs to the true label while maximizing the confidence score of $x$ which belongs to the wrong class in order to achieve the false classification, i.e. targeted evasion. To generate FGM attacks for LS-GNA based NR system, the gradient of the objective function $L\left ( x, \cdot  \right )$ with respect to $x$ is expressed as:
\begin{equation}
\label{equ:gradient of obejective function}
\triangledown L(x,\cdot )=\triangledown S_{y}(x)-\triangledown S_{t}(x)
\end{equation}
To calculate $\triangledown S(x)$,  we first calculate the gradient of the feature vector $f(x)$ of VT-CNN2 classifier to the input data $\frac{\partial f(x)}{\partial x}$ using automatic differentiation package and calculate the gradient of RBF-SVM output with respect to the feature vector $\frac{\partial S(x)}{\partial f(x)}$ manually. Then we obtain $\triangledown S(x)=\frac{\partial S(x)}{\partial f(x)}\cdot \frac{\partial f(x)}{\partial x}$. The gradient of RBF-SVM  $\frac{\partial S(x)}{\partial f(x)}$ is obtained as follows. Let $f(x)$ be replaced by $\xi$ for notational simplicity. The decision function of RBF-SVM classifier is expressed as:
\begin{equation}
\label{equ:decision function of SVM}
S(\xi)=\sum_{i=1}^{K}\alpha _{i}y_{i}exp(-\gamma \left \| \xi-\xi_{i} \right \|^{2})+b
\end{equation}
where $\xi_{i}$ are the support vectors, $y_{i}$ is the class label, $b$ is a bias term, $\alpha _{i}$ are dual variables and $\gamma$ is an RBF kernel parameter. Thus, the gradient of SVM classifier is:
\begin{equation}
\label{equ:gradient of SVM}
\triangledown S(\xi)=\sum_{i=1}^{K}-2\gamma\alpha _{i}y_{i}exp(-\gamma \left \| \xi-\xi_{i} \right \|^{2})\cdot (\xi-\xi_{i})
\end{equation}
Finally, the gradient for generating FGM attack is obtained by combining the  above two gradients using the chain rule.

\begin{algorithm}[ht!]
\caption{FGM-based Adversarial Examples for LS-GNA based NR}\label{alg:FGM}
\hspace*{\algorithmicindent}\textbf{Input: }
\begin{itemize}[leftmargin=1.1cm]
\item input $x$, true label $y$ and the number of classes $N_{c}$
\item the model $S(\cdot ,\theta )$ and the rejection threshold $\Theta$
\item allowed perturbation norm $\varepsilon$, allowed PNR and SNR
\end{itemize}
\hspace*{\algorithmicindent}\textbf{Output: }$\widetilde{x}$: the adversarial examples.

\begin{algorithmic}[1]
\State $\varepsilon =\sqrt{PNR\cdot \left \| x \right \|_{2}^{2}/(SNR+1)}$
\State \textbf{for} $t$ in $range(N_{c})$ \textbf{do}
\State ~~~~$r_{norm}=-1 \cdot (\left \| \triangledown _{x}L(x, e_{t}) \right \|_{2})^{-1}\triangledown _{x}L(x, e_{t})$
\State ~~~~$\widetilde{x} =x+\varepsilon \cdot r_{norm}$

\State \textbf{Until} $arg~max(S(\widetilde{x} ,\theta ))\neq y$ and $max(S(\widetilde{x} ,\theta ))> \Theta $
\State \Return{$\widetilde{x}$}

\end{algorithmic}
\end{algorithm}

\section{Results and Discussion}

All the algorithms are written in PyTorch and executed by NVIDIA GEforce RTX 2080 Ti GPU. The dataset we used is the same as that used in \cite{Sadeghi2019}, i.e., the GNU radio ML dataset RML2016.19a, \href{https://www.deepsig.ai/datasets}{https://www.deepsig.ai/datasets}. It contains 220000 input samples, and each sample corresponds to one modulation scheme at a specific SNR. There are 11 modulation categories in this dataset including BPSK, QPSK, 8PSK, QAM16, QAM64, CPFSK, GFSK, PAM4, WBFM, AM-SSB, and AM-DSB. The samples are generated for 20 different SNR levels from -20dB to 18dB with a step size of 2dB. Each sample has 256 dimensions containing 128 in-phase and 128 quadrature components. Half of the samples are training set and the rest are testing set.

\begin{figure}[ht]
\centering
\includegraphics[scale = 0.54]{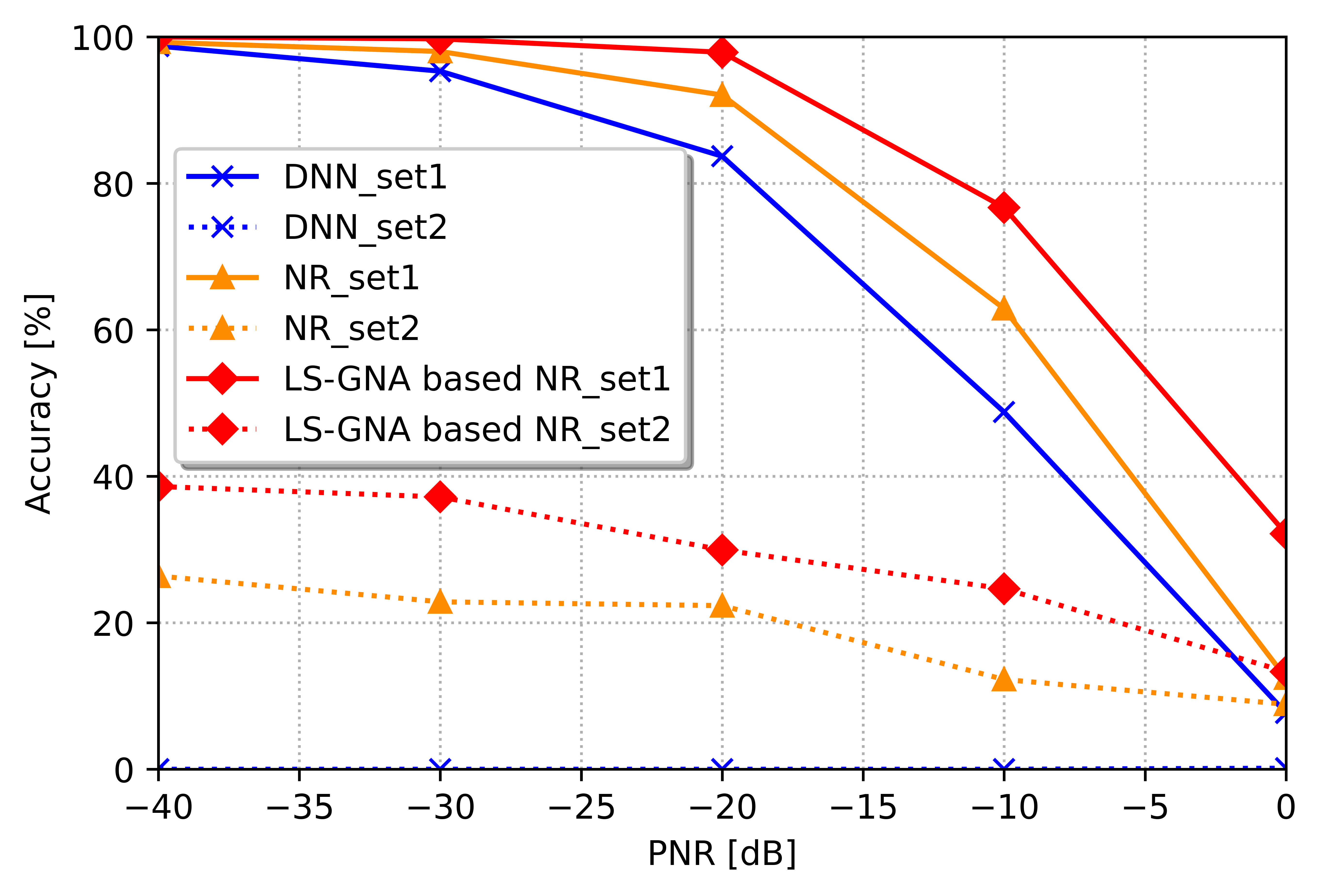}
  \caption{Results of the undefended DNN, the DNN protected with the ordinary NR countermeasure, and the proposed LS-GNA based NR countermeasure.}
  \label{fig:result1}
\end{figure}

\begin{figure}[ht]
\centering
\includegraphics[scale = 0.54]{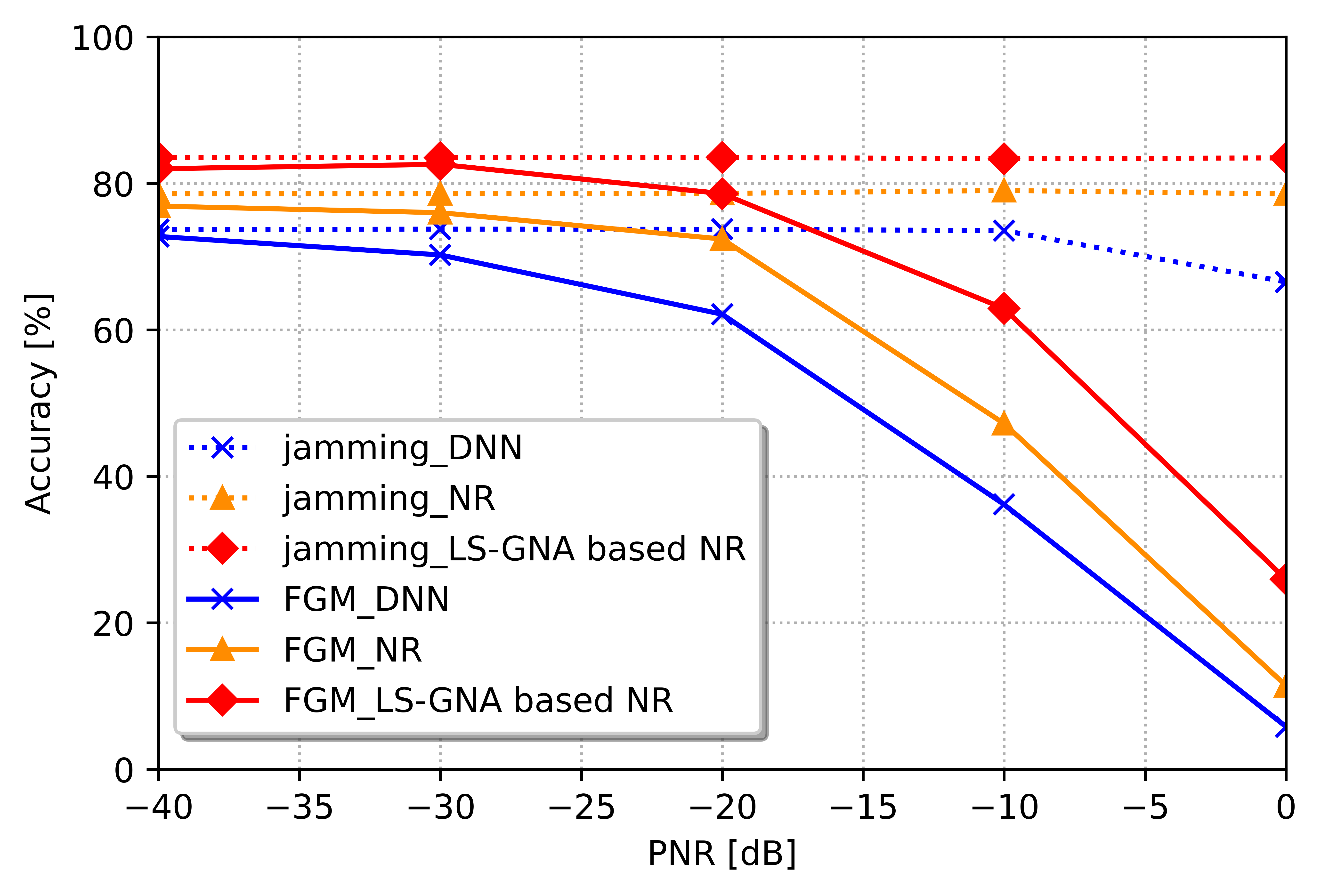}
  \caption{Results of the undefended DNN, the DNN protected with the ordinary NR countermeasure, and the proposed LS-GNA based NR countermeasure against jamming attack.}
  \label{fig:jamming}
\end{figure}

In Figure \ref{fig:result1}, we compare the performance against FGM attacks of the undefended DNN (the same used in \cite{Sadeghi2019}) with the ones of the DNN protected with the ordinary NR countermeasure of \cite{Sotgiu2020}, and the ones of our LS-GNA based NR countermeasure. 
Similar to the case for DNN, we have separated the test data set into set \uppercase\expandafter{\romannumeral1} and set \uppercase\expandafter{\romannumeral2} based on the correct and incorrect classifications by the LS-GNA based NR system. Then the data samples from set \uppercase\expandafter{\romannumeral1} and set \uppercase\expandafter{\romannumeral2} are used to generate adversarial examples respectively. The value of rejection threshold was set so that the rejection rate is 10$\%$ when the original benign data samples from set \uppercase\expandafter{\romannumeral1} are applied. In the absence of adversarial perturbation (i.e., for $\varepsilon = 0$), classification accuracy is computed as usual, but considering rejects as errors; In the presence of adversarial perturbation (i.e., for $\varepsilon > 0$), all test samples become adversarial examples, and we consider them correctly classified if they are assigned either to the rejection class or to their original class (which typically happens when the perturbation is too small to cause a misclassification). It can be observed that for both the datasets \uppercase\expandafter{\romannumeral1} and \uppercase\expandafter{\romannumeral2}, our LS-GNA based NR defense outperforms the unprotected DNN and the simple NR defense. The accuracy of NR defense is also higher than that of the unprotected DNN. We conclude that in terms of the robustness against the FGM attack, both NR and LS-GNA based NR countermeasures outperform a DNN without any countermeasure, i.e., both countermeasures oblige the adversary to use much more perturbation (transmission) power to fool the defender. Furthermore, the performance of DNN, NR, and LS-GNA based NR countermeasure against jamming attacks is compared with that of FGM attacks for the complete dataset (i.e., set I and set II) in Figure \ref{fig:jamming}. In our jamming attacks, the normal signals are augmented with the random Gaussian noise. As seen in Figure \ref{fig:jamming}, the proposed NR and LS-GNA based NR countermeasures are robust against jamming attacks, in addition to its superior performance against FGM based adversarial attacks. Finally, we provide the accuracy (for the complete dataset) of the proposed LS-GNA based NR system against FGM attacks for different modulation schemes in Table \ref{tab:different modulation scheme}. Due to low number of samples for each modulations, we have averaged accuracy performance using 10 repetitions of the training of DNN and LS-GNA based NR system. We present the performance of modulation schemes for which the normal accuracy (i.e., the accuracy without perturbation) is higher than 40$\%$. This is because when the normal accuracy of certain modulation scheme is very low, most of the samples are wrongly classified and can be regarded as adversarial examples even without perturbation. Hence, it is more appropriate to analyze the performance of modulation schemes for which the normal accuracy is high. We observe from Table \ref{tab:different modulation scheme} that the proposed LS-GNA based NR system can defend all the modulations better than the undefended DNN except for PAM4 scheme when $PNR = -10dB$. For all other modulation schemes, LS-GNA based NR outperforms the undefended DNN significantly.

\vspace{-5mm}

\begin{table}[]
\caption{Accuracy of LS-GNA based NR system against FGM attacks for different modulation schemes.}
\label{tab:different modulation scheme}
\begin{center}
\scalebox{0.68}{
\begin{tabular}{ccccc}
\toprule[1pt]
\begin{tabular}[c]{@{}c@{}}Modulation \\ Scheme\end{tabular} & \begin{tabular}[c]{@{}c@{}}DNN\\ (PNR=-10dB)\end{tabular} & \begin{tabular}[c]{@{}c@{}}LS-GNA based NR\\ (PNR=-10dB)\end{tabular} & \begin{tabular}[c]{@{}c@{}}DNN\\ (PNR=0dB)\end{tabular} & \begin{tabular}[c]{@{}c@{}}LS-GNA based NR\\ (PNR=0dB)\end{tabular} \\
\hline
\hline

8PSK                                                         & 1.23                                                          & \textbf{54.35}                                                            & 0.0                                                         & \textbf{5.94}                                                             \\
AM-DSB                                                       & 61.03                                                         & \textbf{85.91}                                                            & 21.01                                                       & \textbf{61.26}                                                            \\
AM-SSB                                                       & 16.48                                                         & \textbf{98.49}                                                            & 0.12                                                        & \textbf{59.22}                                                            \\
BPSK                                                         & 64.41                                                         & \textbf{65.51}                                                            & 0.18                                                        & \textbf{6.96}                                                             \\
CPFSK                                                        & 65.66                                                         & \textbf{97.83}                                                            & 1.57                                                        & \textbf{34.46}                                                            \\
GFSK                                                         & 74.46                                                         & \textbf{89.46}                                                            & 1.33                                                        & \textbf{36.27}                                                            \\
PAM4                                                         & \textbf{86.33}                                                & 82.85                                                                     & 36.97                                                       & \textbf{46.23}                                                            \\
QAM64                                                        & 12.67                                                         & \textbf{59.05}                                                            & 0.35                                                        & \textbf{34.44}                                                            \\
QPSK                                                         & 1.91                                                          & \textbf{31.0}                                                             & 0.0                                                         & \textbf{2.11}                                                            

\\
\toprule[1pt]                                                     
\end{tabular}}
\end{center}
\end{table}

\section{Conclusions}

We have proposed for the first time a defense mechanism against adversarial attacks for automatic modulation classification. Using real radio signals, we have shown that our countermeasure based on a rejection technique, augmented by label smoothing and Gaussian noise injection, has the capability to defend DNN networks against adversarial perturbations. As a consequence of this, adversaries will be forced to increase perturbation (transmission) power to fool the defender. The proposed countermeasures are applicable to the jamming attack scenarios as well. Our future work will extend these techniques to black box and grey box attacks. 

\section*{Acknowledgments}
The authors acknowledge the support of the UK EPSRC through grants EP/R006385/1, EP/N007840/1 and International Scientific Partnership Programme (ISPP-18-134(2)) of King Saud University.

\bibliographystyle{IEEEtran}
\bibliography{references}

\begin{thebibliography}{1}
\providecommand{\url}[1]{#1}
\csname url@samestyle\endcsname
\providecommand{\newblock}{\relax}
\providecommand{\bibinfo}[2]{#2}
\providecommand{\BIBentrySTDinterwordspacing}{\spaceskip=0pt\relax}
\providecommand{\BIBentryALTinterwordstretchfactor}{4}
\providecommand{\BIBentryALTinterwordspacing}{\spaceskip=\fontdimen2\font plus
\BIBentryALTinterwordstretchfactor\fontdimen3\font minus \fontdimen4\font\relax}
\providecommand{\BIBforeignlanguage}[2]{{%
\expandafter\ifx\csname l@#1\endcsname\relax
\typeout{** WARNING: IEEEtran.bst: No hyphenation pattern has been}%
\typeout{** loaded for the language `#1'. Using the pattern for}%
\typeout{** the default language instead.}%
\else
\language=\csname l@#1\endcsname
\fi
#2}}
\providecommand{\BIBdecl}{\relax}
\BIBdecl

\bibitem{6336689}
M.~{Bkassiny}, Y.~{Li}, and S.~K. {Jayaweera}, ``A survey on machine-learning techniques in cognitive radios,'' \emph{IEEE Communications Surveys Tutorials}, vol.~15, no.~3, pp. 1136--1159, 2013.

\bibitem{Swami2000}
A.~Swami and B.~M. Sadler, ``{Hierarchical digital modulation classification using cumulants},'' \emph{IEEE Transactions on Communications}, vol.~48, no.~3, pp. 416--429, 2000.

\bibitem{OShea2018}
T.~J. O'Shea, T.~Roy, and T.~C. Clancy, ``{Over-the-Air Deep Learning Based Radio Signal Classification},'' \emph{IEEE Journal on Selected Topics in Signal Processing}, vol.~12, no.~1, pp. 168--179, 2018.

\bibitem{Scholl2019}
\BIBentryALTinterwordspacing
S.~Scholl, ``{Classification of Radio Signals and HF Transmission Modes with Deep Learning},'' pp. 1--4, 2019. [Online]. Available: \url{https://arxiv.org/abs/1906.04459}
\BIBentrySTDinterwordspacing

\bibitem{goodfellow2014explaining}
I.~J. Goodfellow, J.~Shlens, and C.~Szegedy, ``Explaining and harnessing adversarial examples,'' \emph{arXiv preprint arXiv:1412.6572}, 2014.

\bibitem{Sotgiu2020}
A.~Sotgiu, A.~Demontis, M.~Melis, B.~Biggio, G.~Fumera, X.~Feng, and F.~Roli, ``Deep neural rejection against adversarial examples,'' \emph{EURASIP Journal on Information Security}, no.~5, 2020.

\bibitem{Sadeghi2019}
M.~Sadeghi and E.~G. Larsson, ``{Adversarial attacks on deep-learning based radio signal classification},'' \emph{IEEE Wireless Communications Letters}, vol.~8, no.~1, pp. 213--216, 2019.

\bibitem{Shafahi2019}
\BIBentryALTinterwordspacing
A.~Shafahi, A.~Ghiasi, F.~Huang, and T.~Goldstein, ``{Label Smoothing and Logit Squeezing: A Replacement for Adversarial Training?}'' pp. 1--19, 2019. [Online]. Available: \url{http://arxiv.org/abs/1910.11585}
\BIBentrySTDinterwordspacing

\end{thebibliography}

\end{document}